\documentclass[preprint,12pt, a4paper]{elsarticle}
\usepackage{amssymb}
\usepackage{hyperref}
\usepackage{listings}
\usepackage{caption}
\usepackage{tabularx}

\journal{SoftwareX journal}

\begin{document}

\begin{frontmatter}



\title{SDRDPy: An application to graphically visualize the knowledge obtained with supervised descriptive rule algorithms}


\author[a]{M.A. Padilla-Rascon}
\author[b]{P. Gonzalez}
\author[b]{C.J. Carmona}

\address[a]{Department of Computer Science, University of Jaén, E-23071 Jaen (Spain). eMail: mprascon@ujaen.es}
\address[b]{Andalusian Research Institute in Data Science and Computational Intelligence, University of Jaén, E-23071 Jaen (Spain). eMail: pglez@ujaen.es, ccarmona@ujaen.es}

\begin{abstract} 

SDRDPy is a desktop application that allows experts an intuitive graphic and tabular representation of the knowledge extracted by any supervised descriptive rule discovery algorithm. The application is able to provide an analysis of the data showing the relevant information of the data set and the relationship between the rules, data and the quality measures associated for each rule regardless of the tool where algorithm has been executed. All of the information is presented in a user-friendly application in order to facilitate expert analysis and also the exportation of reports in different formats.
\end{abstract}


\end{frontmatter}

%\linenumbers

%\section*{Comentarios de uan revisión de esta revista}
\section{Motivation and significance}

Currently, the Data Science discipline is booming due to the fact that the amount of information collected digitally is continuously growing and, thanks to Data Science relevant knowledge can be extracted from this information to make decisions \cite{DS2}. Data Science involves several phases \cite{DS1}:%as can be observed in Fig. \ref{fig:phases}:

\begin{enumerate}
\item Data collection. In this phase, all the information from different sources from which knowledge is to be obtained is collected.
\item Selection, cleaning and transformation. In this phase, the data collected in the previous step goes through a filtering process where the variables to be studied are selected, and incomplete or erroneous data are eliminated and transformed so that the Data Science technique can process this information.
\item Data mining. In this phase, it is decided which algorithm is going to be applied to the data set depending on the technique to be applied (supervised or unsupervised learning, descriptive or predictive, etc) and the way in which the results are to be obtained (set of rules, decision trees, etc.).
\item Evaluation. The results obtained in the previous phase are evaluated by experts in order to determine the quality of the resulting model and whether it is necessary to repeat the experiment modifying certain parameters to improve it or to observe other aspects.
\item Dissemination and use. Once the desired models and the knowledge implicit in the collected information have been obtained, it is necessary to present this information in a user-friendly and easy-to-understand way.
\end{enumerate}

Data scientists have been specializing for each of the stages described previously, so that Data Science is as optimized as possible and the best possible results are found. However, there is a lack in the phase related to the dissemination and use of the knowledge because the same amount of work and time has not been invested in creating interfaces that are usable and user-friendly. Definitely, experts have tools that are not capable of interpreting the results obtained in the different experimental studies and they need to be assisted by data scientist in order to understand the studies.

This problem also occurs in supervised descriptive rule discovery (SDRD) \cite{Cdh18}. This method groups a set of techniques such as Subgroup Discovery \cite{Hcgd11}, Emerging Pattern Mining\cite{Gcmgj18}, Contrast Set Mining \cite{Bp01} and Change Mining \cite{Ccc05} are included. These techniques combine descriptive induction in supervised learning, and they do not attempt to predict/classify new instances but rather describe the main characteristics/descriptors of a problem. They have been used throughout the literature in order to solve problems in different domains like medicine, industry, e-commerce, and so on. In short, experts do not have a tool for visualizing the results of the SDRD algorithms in an intuitive and attractive way, and therefore SDRDPy tries to fill this gap by developing a usable graphical interface that allows any user to consult the results obtained from the executions in a graphical and tabular way, based on the knowledge obtained by the algorithms that have been executed in the software where they were developed, i.e. SDRDPy processes the rules extracted by the algorithms without the need for new executions and/or adaptations of their codes.

\section{Existing applications}
There are several applications that allow the execution and visualization of different types of algorithms, including SDRD algorithms. Some of these applications are the following:

\begin{itemize}
    \item $KEEL$ \footnote{\url{http://www.keel.es}}: It is a free software tool developed in JAVA language. This application does not offer algorithm visualization functionalities, but it allows users to execute the integrated algorithms. It is useful for obtaining the data files in the format compatible with the application.
    \item $Weka$ \footnote{\url{https://www.cs.waikato.ac.nz/ml/weka/}}: It is a free software tool that provides a wide variety of machine learning and data mining algorithms. It allows users to load datasets from various sources, preprocess the data and apply different machine learning algorithms. However, it does not have SDRD algorithms, it can convert an association rule algorithm into SD if a variable is prefixed, but it does not work with SD concepts per se rather with association rules with conditions.
    \item $KNIME$ \footnote{\url{https://www.knime.com/}}: It is an open-source platform for data integration, analysis, and exploration. It provides a graphical interface that allows users to build data workflows by connecting nodes representing various operations, from data loading to advanced analysis and visualization. It also supports the integration of other languages and tools. Nevertheless, similar to Weka, it does not include SDRD algorithms but instead employs association rule algorithms with a predefined rule.
    \item $Orange$ \footnote{\url{https://orangedatamining.com/}}: It is an open source visual data mining and data analysis tool that offers a wide range of machine learning and data mining algorithms. In Orange, there are two supervised descriptive rule (SDRD) algorithms: CN2-SD and AprioriSD. These algorithms have been included through an additional plugin developed by researchers. Orange does not allow the analysis of the data itself, only the analysis of the rules obtained. Users can add and customize visualization widgets using Python, although this feature has the limitation that the algorithms and their output must be adapted to the architecture specified by Orange.
\end{itemize}

The $SDRDPy$ application stands out from the rest because:
\begin{itemize}
    \item It is not necessary that the output file after the execution of the algorithm has to be adapted to the application. Working with the algorithm is independent of the programming language, working framework, tool, etc.
    \item It allows the analysis of both data and rules, and the relationship between them.
    \item Nowadays, it includes six different SDRD algorithms and the inclusion of new algorithms is simple following the manual\footnote{\url {https://github.com/mariasun-pr/SDRDPy/tree/master}}.
\end{itemize}

\section{Software Description}

This section will cover:
\begin{enumerate}
    \item The architecture of the system in section \ref{subsec:architecture}, where the libraries used are indicated and the functioning of the application is explained.
    \item The functionalities of the system in the section \ref{subsec:func}.
\end{enumerate}

\subsection{Software architecture} \label{subsec:architecture}

SDRDPy is a desktop application that has been implemented using Python version 3.11, so it can be used from this version onwards, and it has the following dependencies and libraries:

\begin{itemize}
    \item The Tkinter library was used to create the interface.
    \item Matplotlib facilitated the creation of graphics.
    \item Numpy handled arrays.
    \item Re managed character strings.
    \item Ziplib was utilized to create ZIP files of the reports.
\end{itemize}

\begin{figure*}[!htpb]
	\centering
\includegraphics[width=\linewidth]{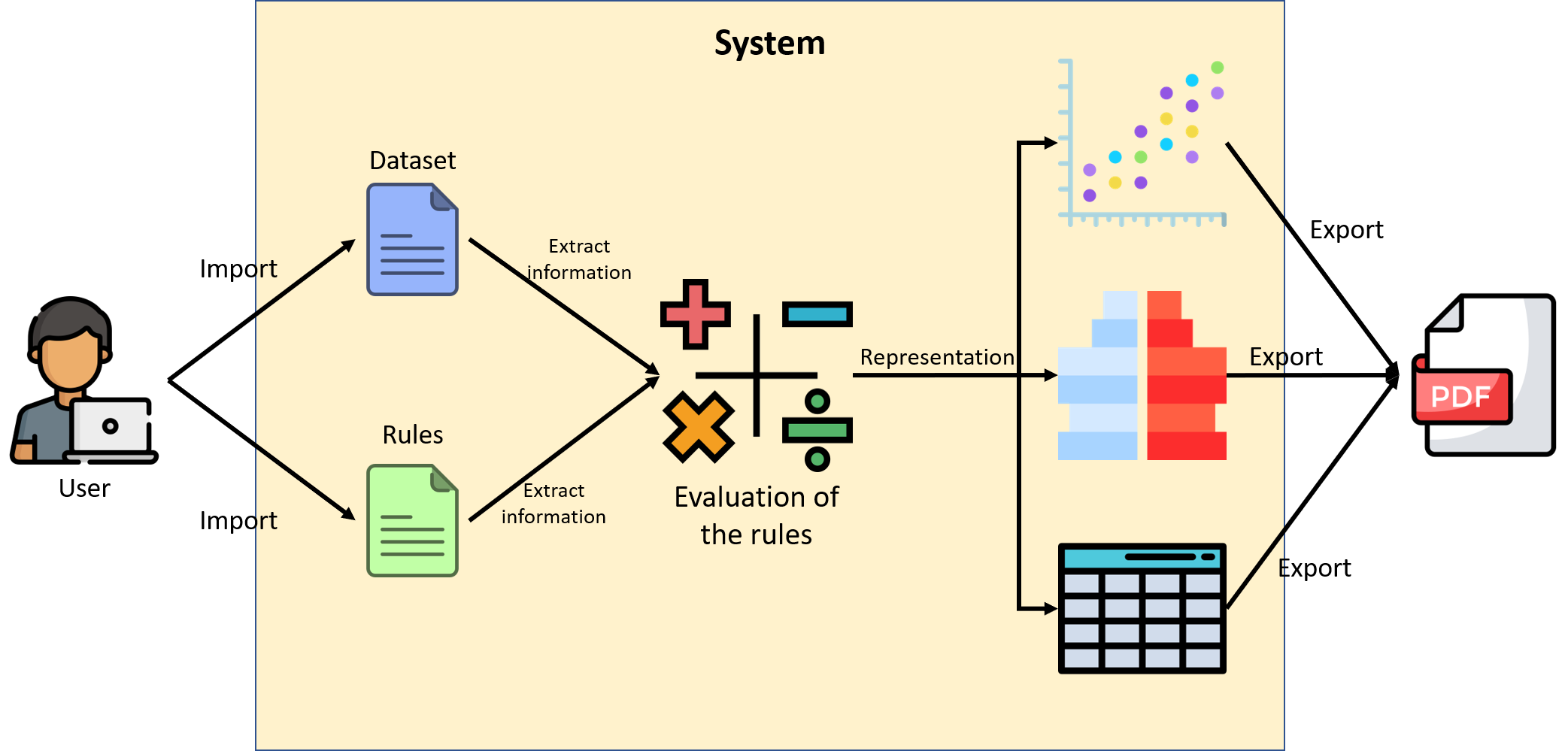}
	\caption{System Architecture.}
	\label{fig:arq}
\end{figure*}

Figure \ref{fig:arq} represents the architecture of SDRDPy where in summary, the system handles all data processing, calculations and visualization internally, allowing the user to focus on interpreting the results and making decisions based on them.

The user needs to import the data file and the rules file in any order, i.e. the user can import the data file first and then the rules file or vice versa in order to use SDRDPy. It is important to note that:

\begin{itemize}
    \item  The data file must contain the data set to be used for the evaluation of the rules. The user can either import the complete file, which will contain both the training set and the test set, or import either the training file or the test file. 
    \item The data set must be in the format of the data sets provided by $KEEL$ \ref{lst:data}, otherwise the application will not detect it correctly. This data format is very similar to the standard Weka format.
    \item Once the data file has been imported, the user must import the rules file to be evaluated. This file must contain the name of the algorithm that generated the rules and the rules to be applied to the data set to obtain the desired results \ref{lst:reglas}.
\end{itemize}

\lstset{
  basicstyle=\small\ttfamily,
  breaklines=true,
  columns=fullflexible,
  captionpos=b,
  frame=lines,
}
\begin{lstlisting}[label={lst:data},caption={Header and format of the data in the data set file.}]
@relation iris
@attribute sepalLength real [4.3, 7.9]
@attribute sepalWidth real [2.0, 4.4]
@attribute petalLength real [1.0, 6.9]
@attribute petalWidth real [0.1, 2.5]
@attribute class {Iris-setosa, Iris-versicolor, Iris-virginica}
@inputs sepalLength, sepalWidth, petalLength, petalWidth
@outputs class
@data
5.1, 3.5, 1.4, 0.2, Iris-setosa
4.9, 3.0, 1.4, 0.2, Iris-setosa
\end{lstlisting}

At this point, SDRDPy internally extracts the necessary data and performs the corresponding calculations to evaluate the rules. The representation of the supervised descriptive rules obtained always has the same form. This feature allows the definition of the same contingency table \ref{tab:tableCon} for any rule ($R$). The calculation of this table is fundamental because it is possible to obtain several quality measures that allow us to evaluate the effectiveness and usefulness of the rule. A detailed review of the measures can be found in the paper and SDRDPy considers the following quality measures \cite{Gcmgj18}:

\begin{lstlisting}[label={lst:reglas},caption={Header and format of the rules in the rules file generated by difuse algorithms.}]
@algorithm nmeef
Number of labels: 3
GENERATED RULE 0
    Antecedent
        Variable petalLength = Label 0 	 (-1.95 1.0 3.95)
    Consecuent: Iris-setosa
\end{lstlisting}

\begin{itemize}
    \item TPr (True Positive rate): It is the ratio of true positives to the total number of examples belonging to the class.
    \item FPr (False Positive rate): It is the ratio of false positives to the number of examples that do not belong to the class.
    \item Conf (Confidence): Measures the accuracy of the rule with respect to the examples it covers.
    \item WRAcc (Unusualness): Measures the balance between generality and precision. A normalised version of this measure is used in the application \cite{Cdh18}.
\end{itemize}

\begin{table}
    \centering
        \begin{tabularx}{\columnwidth}{X|X|X|X}
        & Positives (Class examples) & Negatives (Examples of non class) & \\
        \hline\hline
        Covered & $p=tp$ & $n=fp$ & $p+n$ \\
        \hline
        Not covered & $\overline{p} = fn$ & $\overline{n}=tn$ & $\overline{p}+\overline{n}$ \\
        \hline
        & $p+\overline{p}=P$ & $n+\overline{n}=N$ & $P+N=T$ \\
        \end{tabularx}
    \caption{Contingency table}
    \label{tab:tableCon}
\end{table}

Finally, the system generates the graphical representations of the results so that the user can clearly visualize them. Moreover, users can also request that the system generates an export file containing the information displayed in the application for later use.

\subsection{Software functionalities}\label{subsec:func}

The main purpose of the application is to allow the user to graphically visualize the analysis of any supervised descriptive rule algorithm resulting from the execution on a data set. The main functionalities are:

\begin{itemize}
    \item Visualize graphically the general analysis of all the rules.
    \item Visualize the contingency table for each of the rules.
    \item Visualize graphs showing the results recorded in the contingency table.
    \item Visualize the table of quality measures (Confidence, WRAcc, TPr and FPr) for each of the rules.
    \item Show the rules that cover each data.
    \item Show the data covered by a rule.
    \item Export to PDF the information displayed on screen.
\end{itemize}

\section{Illustrative Examples}

The application is able to analyze the results obtained from both fuzzy and crisp SDRD algorithms. Although, the evaluation has some differences between both, specifically, in fuzzy algorithms:

\begin{itemize}
    \item The number of labels considered for variable with continuous domains must be indicated in the rule file.
    \item It is necessary to calculate the degree of membership of linguistic labels to the variables, and to the rule, definitely.
    \item The degree of membership is shown in the visualization, but not in crisp algorithms because they do not have it.
\end{itemize}

An illustrative example of the visualization of the supervised descriptive rules obtained in the execution of the NMEEFSD \cite{nmeef} algorithm and APRIORISD algorithm \cite{apriori}, both with the iris dataset \footnote{\url{https://www.keel.es/datasets.php}} is analyzed in this section. 

NMEEFSD is a multiobjective evolutionary algorithm designed to address the problem of subgroup discovery. Its main objective is to find meaningful subgroups of data that are associated with multiple targets or criteria of interest, rather than just one. Throughout the literature, this algorithm has shown significant interest in multiple real-world domains \cite{Cstbdg12}\cite{Ccsd13}\cite{Cgerg23}. This algorithm has been considered due to its interest in the literature, and because it provides the main difference with respect to the crisp algorithms. It adds the degree of triggering of the rule in the application. 

Next, the visualization of the results obtained from the execution of this algorithm is shown below. Section \ref{subsec:gen} displays the most important parts of the general rule information, and Section \ref{subsec:conc} presents the key parts of a particular rule's information.

\subsection{General Visualization of Rules}\label{subsec:gen}

\begin{figure*}[!hbtp]
	\centering
\includegraphics[width=\linewidth]{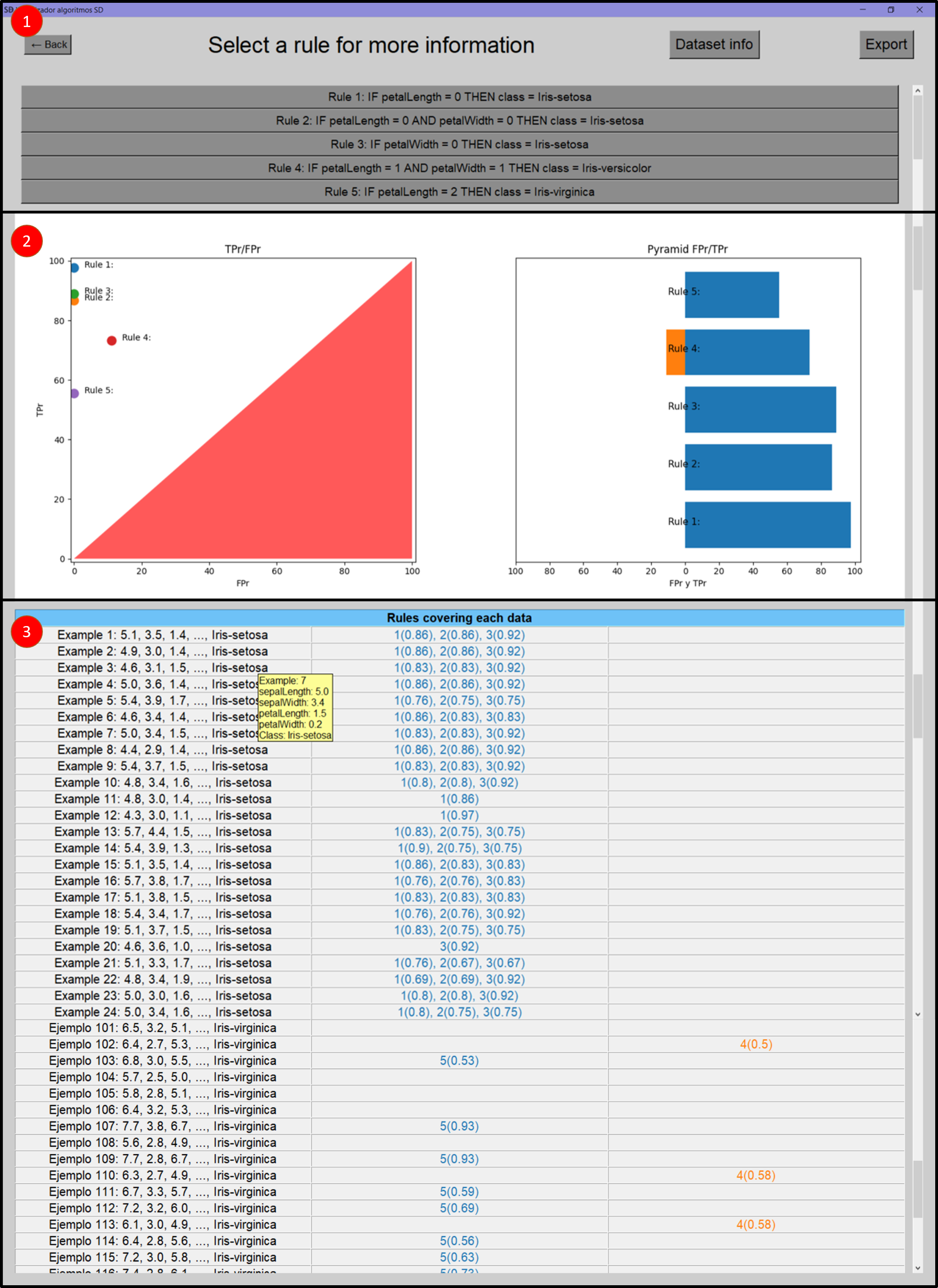}
	\caption{Overview of general info (NMEEF algorithm).}
	\label{fig:pantallaGeneral}
\end{figure*}

\begin{figure*}[!hbtp]
	\centering
\includegraphics[width=\linewidth]{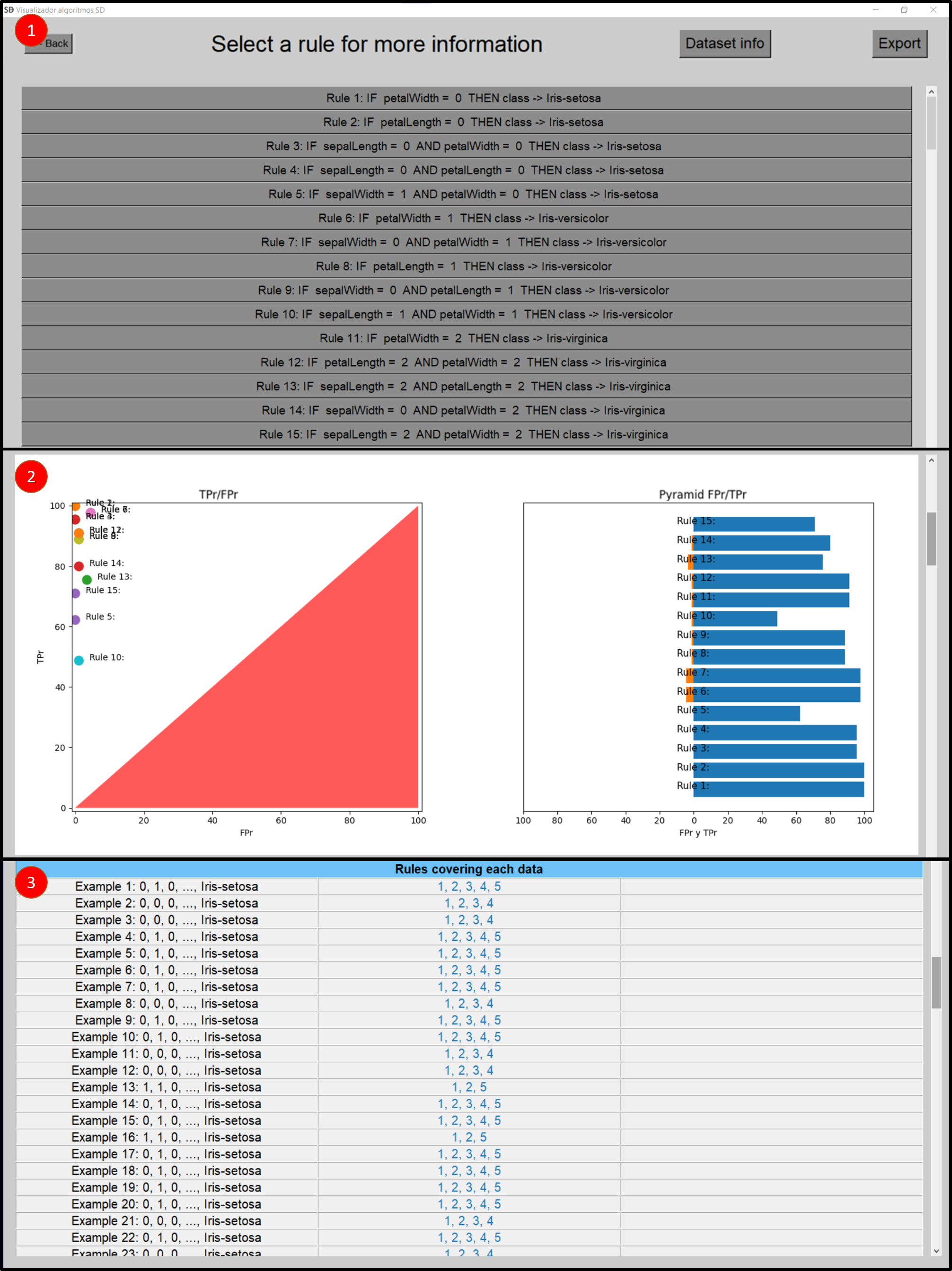}
	\caption{Overview of general info (APRIORISD algorithm).}
	\label{fig:pantallaGeneralA}
\end{figure*}

Once the application has internally evaluated the rules, it displays a screen with three distinct visualizations as can be observed in Fig. \ref{fig:pantallaGeneral} (NMEEFSD algorithm) and Fig. \ref{fig:pantallaGeneralA} (APRIORISD algorithm).

In the first part, all the rules contained in the rules file are displayed (one per row), allowing the user to consult the information of each of them individually if the user clicks on any of them. This action will take the user to section \ref{subsec:conc}.

The second part is composed of graphs. The one on the left is a dot plot showing the generalized dispersion of the rules according to their TPr and FPr following the recommendations of this contribution \cite{Cdh18}. On the other hand, on the right is a pyramid plot showing the distribution of TPr and FPr for each of the rules \cite{Nlw09}.

Finally, the third part of the display consists of a table with three columns. The first column shows the corresponding data, the second column in blue shows the rules that have correctly covered that data and the third column in orange shows the rules that have wrongly covered that data. All the rules have the degree of belonging to the data in brackets to the right of them. If the user holds the cursor over the data, the attributes of the data are expanded. This part is especially useful for understanding how the rules are classifying the data and to identify patterns and trends in the classification.

\subsection{Visualization of a specific rule}\label{subsec:conc}

\begin{figure*}[!hbtp]
	\centering
\includegraphics[width=\linewidth]{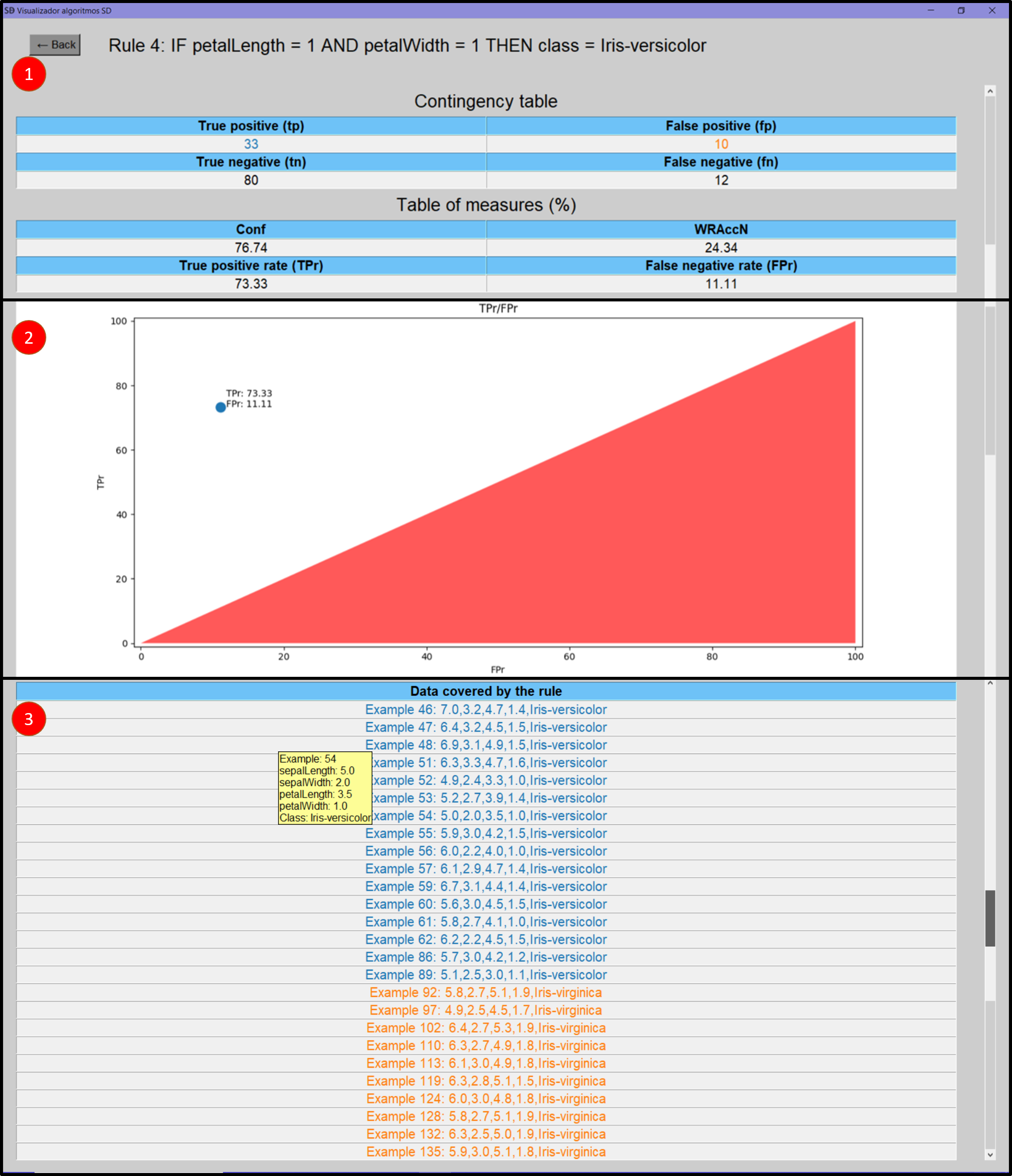}
	\caption{Overview of a specific rule (NMEEFSD algorithm).}
	\label{fig:generalRegla}
\end{figure*}

\begin{figure*}[!hbtp]
	\centering
\includegraphics[width=\linewidth]{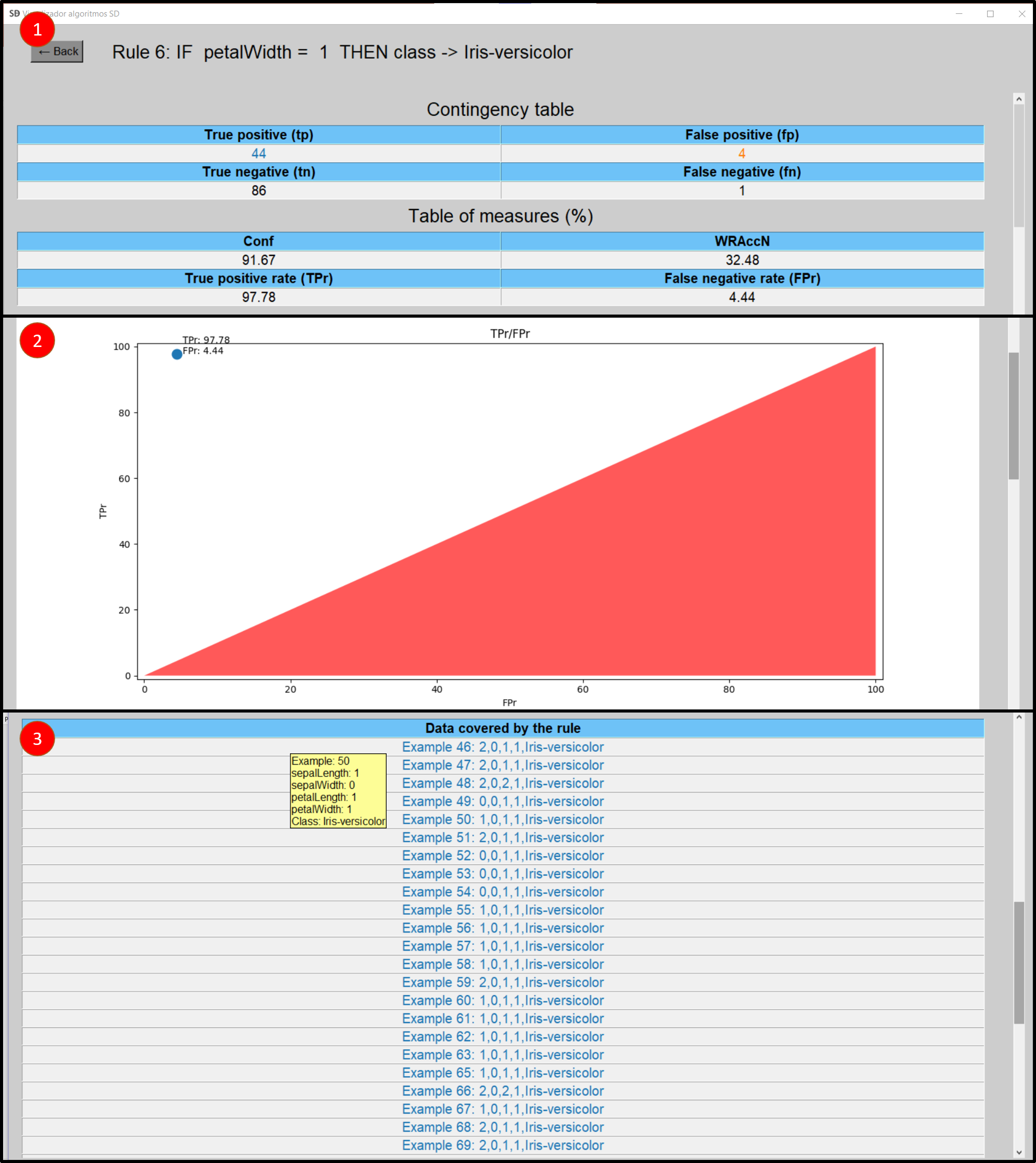}
	\caption{Overview of a specific rule (APRIORISD algorithm).}
	\label{fig:generalReglaA}
\end{figure*}

The application is also able to show information for each rule in order to facilitate its analysis to the experts. The specific information of a rule is accessed by clicking on the name of the rule (as we have mentioned in the previous section). The application shows a new screen where three different parts are highlighted, as can be observed in Fig. \ref{fig:generalRegla} (NMEEFSD algorithm) and Fig. \ref{fig:generalReglaA} (APRIORISD algorithm).

In the first part the user will find the tables with the quality measurements of the rules: 
\begin{itemize}
    \item The contingency table indicates how many examples the rule covered and how many it did not cover, as well as how many it covered correctly (in blue) and how many it covered incorrectly (in orange).
    \item The measures table represents the evaluation measures, including Confidence, WRAcc, TPr and FPr. 
\end{itemize}

The second part presents a dot plot that allows visualizing the dispersion of the rule from the TPr versus FPr. This plot is particularly useful for analyzing the performance of the rule with respect to their interest. To highlight the worst quality rules, a red area has been drawn on the x=y line, as the rules in this area have a lower TPr than the FPr, indicating that the rules are of very poor quality. On the other hand, a rule will be better the closer it is to the (0,100) coordinate, i.e. the closer it is to the upper left corner.

The third section is a table that lists all the data that the rule has covered, indicating whether it has covered them correctly or not. Data covered correctly are shown in blue, while data covered incorrectly are shown in orange. This table provides detailed information on how the rule is classifying the data and allows us to identify possible errors in the classification process.

\section{Impact}

SDRDPy is a particularly useful application for experts focused on the study of SDRD algorithms. This application allows them to visualize the results obtained after the execution of these algorithms. This application offers an analysis of the data, showing the relevant information of the data set and the relationship between the rules and the data. 

Importantly, the application adapts to any algorithm without the researcher having to adapt it for the application to detect it, therefore, it can be expanded to more algorithms within SDRD. Moreover, its versatility extends to complex environments such as Big Data and Data Streaming, allowing users to work with large volumes of data.

Finally, it is important to remark that this application also offers the possibility of exporting the information to different formats, allowing users to extract the information they need, facilitating the analysis and dissemination of the results obtained with the algorithm.

\section{Conclusions}

SDRDPy, is not an exclusive solution in its field, given that there are other tools such as Orange with similar objectives. The application developed is distinguished by its outstanding feature: the inclusion of a wide variety of SDRD algorithms. This particularity makes it a specially appropriate tool for the analysis of supervised descriptive rules obtained through these algorithms.

Moreover, SDRDPy is a desktop application that fills a gap in visualization tools for supervised rule discovery (SDRD) algorithms. SDRDPy focuses on providing graphical and tabular information in a clear and concise manner to facilitate the understanding and identification of patterns and trends in the data. In addition to the visualization of the rule information, the application will also allow the user to easily access the information in the data set, as well as export the information displayed in the application.

On the other hand, SDRDPy is under continuous development to increase the number of algorithms it is able to process as it is an easily extendable application. At the same time, it is a user-friendly, intuitive and usable application, indicating to the user what to do and what information to provide at each step.

\section*{Acknowledgments} 

This work is financed by the Ministry of Science, Innovation and Universities with code PID2019-107793GB-I00/AEI/10.13039/501100011033.



%% References:


%\begin{thebibliography}{00}
%%% Text of bibliographic item%
%
%\bibitem{}
%
%\end{thebibliography}

%\bibliographystyle{elsarticle-num}
%\bibliography{bibliografia}

\appendix

\section*{Current code version}
\label{}

\begin{table}[!h]
\begin{tabular}{|l|p{2.5cm}|p{4.5cm}|}
\hline
\textbf{Nr.} & \textbf{Code metadata description} & \textbf{Please fill in this column} \\
\hline
C1 & Current code version & v5 \\
\hline
C2 & Permanent link to code/repository used for this code version & \url{https://github.com/mariasun-pr/SDRDPy/tree/master} \\
\hline
C3  & Permanent link to Reproducible Capsule & \url{https://github.com/mariasun-pr/SDRDPy/tree/master}\\ 
\hline
C4 & Legal Code License   & None \\
\hline
C5 & Code versioning system used & git \\
\hline
C6 & Software code languages, tools, and services used & Python \\
\hline
C7 & Compilation requirements, operating environments \& dependencies & Python 3.11, Matplotlib and Numpy.\\
\hline
C8 & If available Link to developer documentation/manual & \url{https://github.com/mariasun-pr/SDRDPy/tree/master/Manuals/} \\
\hline
C9 & Support email for questions & mprascon@ujaen.es\\
\hline
\end{tabular}
\caption{Code metadata}
\label{} 
\end{table}

\section*{Current executable software version}
\label{}

\begin{table}[!h]
\begin{tabular}{|l|p{2.5cm}|p{4.5cm}|}
\hline
\textbf{Nr.} & \textbf{(Executable) software metadata description} & \textbf{Please fill in this column} \\
\hline
S1 & Current software version & Version 5.0 \\
\hline
S2 & Permanent link to executables of this version  & \url{https://github.com/mariasun-pr/SDRDPy/tree/master/Executable%20Windows} \\
\hline
S3  & Permanent link to Reproducible Capsule & \url{https://github.com/mariasun-pr/SDRDPy}\\
\hline
S4 & Legal Software License & None \\
\hline
S5 & Computing platforms/Operating Systems & Using by console: MacOS, Windows and Linux.

Using by executable: Windows. \\
\hline
S6 & Installation requirements \& dependencies & To use the application by console is necessary have Python 3.11, Matplotlib and Numpy.\\
\hline
S7 & If available, link to user manual - if formally published include a reference to the publication in the reference list & \url{https://github.com/mariasun-pr/SDRDPy/blob/master/Manuals/User_Guide.md}\\
\hline
S8 & Support email for questions & mprascon@ujaen.es\\
\hline
\end{tabular}
\caption{Software metadata}
\label{} 
\end{table}

\end{document}